\documentclass[letterpaper]{article} 
\usepackage{aaai25}  
\usepackage{times}  
\usepackage{helvet}  
\usepackage{courier}  
\usepackage[hyphens]{url}  
\usepackage{graphicx} 
\usepackage{natbib}  
\usepackage{caption} 
\usepackage{algorithm}
\usepackage{algpseudocode}
\usepackage{subcaption}
\usepackage{amsmath}
\usepackage{amssymb}

\usepackage{newfloat}
\usepackage{listings}
\DeclareCaptionStyle{ruled}{labelfont=normalfont,labelsep=colon,strut=off} 
\floatstyle{ruled}
\newfloat{listing}{tb}{lst}{}
\floatname{listing}{Listing}
\title{PIXELS: \underline{P}rogressive \underline{I}mage \underline{X}emplar-based \underline{E}diting with \underline{L}atent \underline{S}urgery}
\author{
    Shristi Das Biswas\textsuperscript{\rm 1},
    Matthew Shreve\textsuperscript{\rm 2},
    Xuelu Li\textsuperscript{\rm 2},
    Prateek Singhal\textsuperscript{\rm 2},
    Kaushik Roy\textsuperscript{\rm 1}
}
\affiliations{
    \textsuperscript{\rm 1}Purdue University,
    \textsuperscript{\rm 2}Amazon Fashion\\
    sdasbisw@purdue.edu,
    \{mashreve, xueluli, prtksngh\}@amazon.com,
    kaushik@purdue.edu

}
\usepackage{bibentry}
\begin{document}
\twocolumn[{
\renewcommand\twocolumn[1][]{#1}
\maketitle
\begin{center}
    \centering 
    \vspace{-5pt}
    \includegraphics[width=0.93\textwidth]{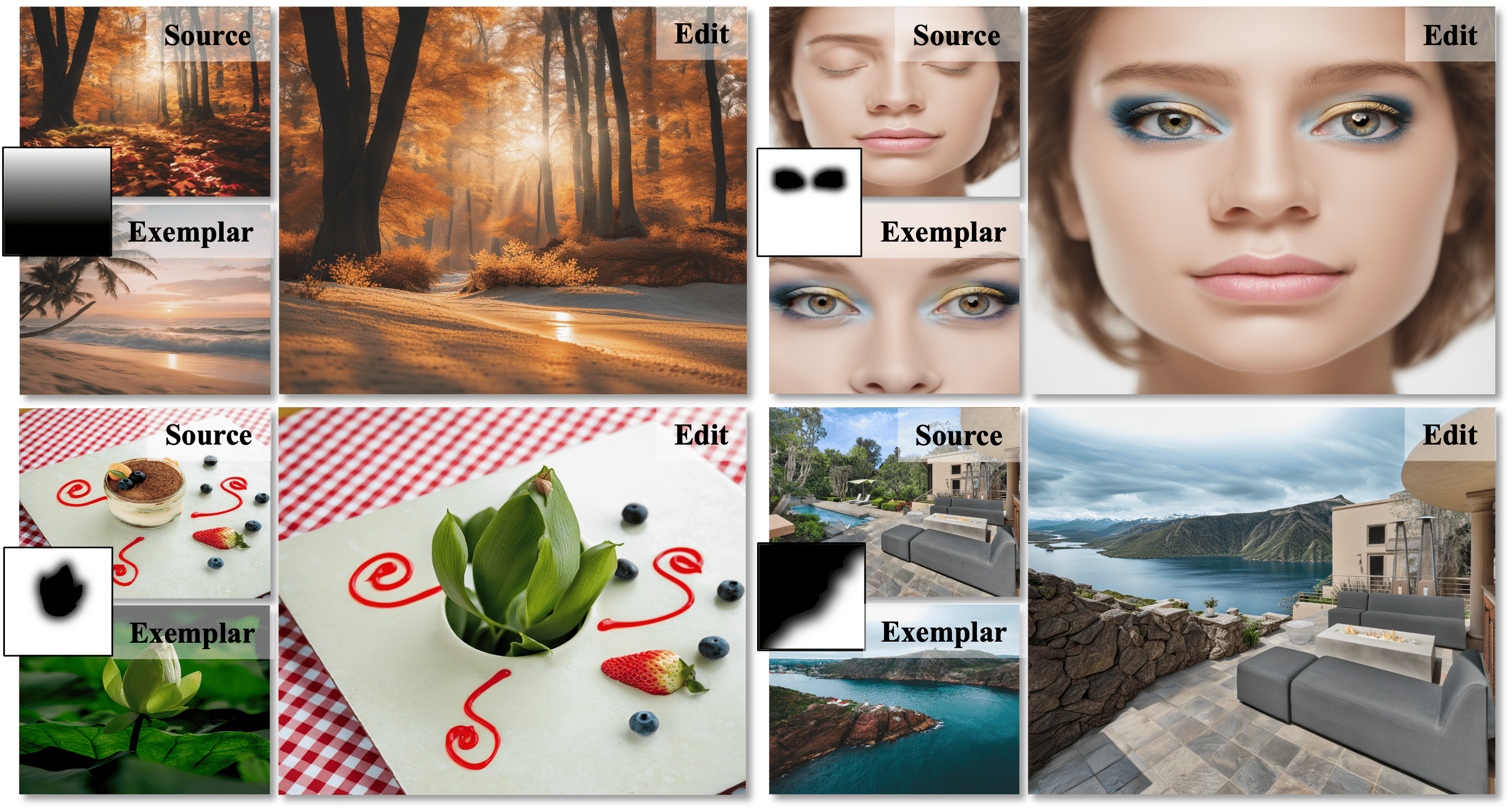}
    \vspace{-5pt}
    \captionof{figure}{\textbf{In-the-wild editing results} produced by our method where users specify to-edit regions in the source image along with exemplars to inspire the edit. Our method changes different regions of the source by different amounts, according to a given non-binary edit map: the darker the region; more flexibility to adapt to the exemplar. This controllability allows us to create gradual spatial changes (e.g., forest-to-beach transition, top left) and transition across an edit realistically.}
    \label{fig:teaser}
\end{center} 

}] 
\normalsize
\begin{abstract}
Recent advancements in language-guided diffusion models for image editing are often bottle-necked by cumbersome prompt engineering to precisely articulate desired changes. An intuitive alternative calls on guidance from in-the-wild image exemplars to help users bring their imagined edits to life. Contemporary exemplar-based editing methods shy away from leveraging the rich latent space learnt by pre-existing large text-to-image (TTI) models and fall back on training with curated objective functions to achieve the task. Though somewhat effective, this demands significant computational resources and lacks compatibility with diverse base models and arbitrary exemplar count. On further investigation, we also find that these techniques restrict user control to only applying uniform global changes over the entire edited region.
In this paper, we introduce a novel framework for progressive exemplar-driven editing with off-the-shelf diffusion models, dubbed PIXELS, to enable customization by providing granular control over edits, allowing adjustments at the pixel or region level. Our method operates solely during inference to facilitate imitative editing, enabling users to draw inspiration from a dynamic number of reference images, or multimodal prompts, and progressively incorporate all the desired changes without retraining or fine-tuning existing TTI models. This capability of fine-grained control opens up a range of new possibilities, including selective modification of individual objects and specifying gradual spatial changes. We demonstrate that PIXELS delivers high-quality edits efficiently, leading to a notable improvement in quantitative metrics as well as human evaluation. By making high-quality image editing more accessible, PIXELS has the potential to enable professional-grade edits to a wider audience with the ease of using any open-source image generation model. 
\end{abstract}


\begin{links}
    \link{Code}{https://github.com/amazon-science/PIXELS}
\end{links}

\section{Introduction}
Language-guided image generation using large text-to-image (TTI) diffusion models trained on web-scale data has made remarkable strides in recent years~\cite{dhariwal2021diffusion, ho2020denoising, nichol2021glide, ramesh2022hierarchical, ye2024altdiffusion, balaji2022ediffi, shi2024instantbooth}. As the need for image editing applications for creating novel content became pervasive in the age of social media, pipelines for creative editing have become a popular demand. Powered by these large-scale pre-trained TTI models, editing methods~\cite{chen2024anydoor, cao2023masactrl, mokady2023null} have been proposed that allow manipulating content in images with the guidance of an input text prompt. However, these editing approaches~\cite{brooks2023instructpix2pix, hertz2022prompt} still find it challenging to fit the requirements of complicated practical scenarios. For instance, as shown in Fig.~\ref{fig:teaser}, the method needs to modify the porch just enough to circumvent abrupt transitions when editing with the lake-view exemplar by generating intermediate features that realistically connect them, or even in the case of a creative edit imagining a bud seamlessly replace the given dessert. Flexibility to allow any kind of edits is important for real applications like custom scene design, product creation and placement, special effects, etc. 

Inpainting approaches~\cite{yu2023inpaint, xie2023smartbrush} take a binary mask and text guidance to regenerate the masked region using TTI models. However, they are limited by the complexity of prompt engineering to accurately reflect user-desired effects, since detailed object appearance features often cannot be feasibly specified by plain text. To this effect, the need for a ``universal language'' that enables precise and intuitive control has become ubiquitous. To solve these shortcomings, there have been efforts to drive editing through the use of exemplars~\cite{song2024imprint, yuan2023customnet, chen2024zero, chen2024anydoor} which take as input a binary mask to introduce the main subject from an exemplar into the source image. However, such approaches often struggle to deal with local edits and are strictly restricted to using only one exemplar at a time. In addition, they require pairs of object masks from video frames for their training, which are difficult to obtain at scale and restrict their data diversity during training. Other exemplar-based methods~\cite{yang2023paint, xu2023versatile} suffer from similar limitations where they train specialized architectures to learn semantic editing of single objects into an image. We believe that instead of encouraging this trend of training custom pipelines on limited datasets, initiatives should be taken towards ``reusing'' the information learnt by pre-existing image generation models that have already been trained on much larger datasets.

To satisfy these requirements, we propose a method to improve exemplar-based editing to exercise creativity more freely. Our main contributions are: (1) We allow users to define the `editability' of each pixel in a picture by different strengths efficiently and simultaneously for seamless edits using a dynamic number of in-the-wild reference images. This is mainly achieved by the insight that selectively modifying various regions at different timesteps during a generation model's inference process induces control over their fidelity to the image on a spatial basis. When introducing an exemplar into the source, say for creating a new image that replaces the leaf-strewn ground with a sandy beach (Fig.~\ref{fig:teaser} top-left), we would like to introduce different amounts of changes into the edited regions on the photo, in a controllable manner. This way, source image pixels closer to the edit boundary attain more flexibility to transition to the new exemplar concept harmoniously. This enables finer granularity across an edit, opening ways the user can exert more control over changes than is attainable using previous methods. (2) We show how to extend our algorithm to introduce any number of exemplars into an image, making it a more suitable candidate for general-purpose editing in contrast to the exemplar count restrictions imposed by other models. (3) Our framework only requires adapting the inference process of existing off-the-shelf image generation models~\cite{rombach2022high, podell2023sdxl} trained on web-scale data, to leverage their rich latent space and pre-learnt knowledge of contextual relationships between objects for realistic edits. This enables a heightened level of refinement to mitigate any artifacts or unnatural transitions for exemplar-driven editing. (4) We do not forsake the original text-guidance capability of these models, leaving it up to the user to exert any form or degree of control. (5) Finally, our approach performs favorably over the prior art for the task, measured by both quantitative metrics and user evaluation.

\section{Related Works}
\hspace{8pt}\textbf{Text-driven image editing:} With recent strides in TTI models, language-guided editing has been largely explored. Traditional approaches~\cite{andonian2021paint, gal2022stylegan, patashnik2021styleclip, xia2021tedigan} fall back on pretrained-GAN generators~\cite{karras2020analyzing} and text encoders~\cite{radford2021learning} to achieve instance-based optimization according to text prompts. However, these approaches are often cost-ineffective due to progressive optimization steps on the image and struggle with complex editing due to the limited modeling capability of GANs. Recently, diffusion models (DMs) have risen as the new state-of-the-art for TTI generation~\cite{song2020denoising, song2020score}. A consecutive line of work~\cite{kim2021diffusionclip, ruiz2023dreambooth, kawar2023imagic} fine-tuned these DMs for each target text. The high computational cost of fine-tuning case-specifically for every image, however, makes them impractical for interactive image editing. Authors in~\cite{avrahami2023blended} perform a sequence of noise-denoise operations to create local edits given a mask. More recent works like~\cite{hertz2022prompt,liu2023more} still suffer from the lack of precise control with text guidance to explicitly convey the user’s imagination.

\textbf{Exemplar-driven image editing:} Parallel to the direction of using text guidance to describe desired edits, a more recent line of work alternatively focuses on guidance from visual exemplars. Early works in using images to guide generation~\cite{meng2021sdedit, seo2023midms} focused on modifying the entire image, making them incompatible for edits with local strength control. Methods such as~\cite{yang2023paint} and~\cite{song2023objectstitch} extract representations from the reference image using a CLIP encoder to inpaint into a source image while~\cite{ye2023ip} trains an image-adapter to achieve visual prompt capability for pre-trained text-to-image models. Later works~\cite{xu2023versatile, chen2024zero} train specialized architectures with curated objective functions to solve the editing task. While they allow exemplar-guidance, the four major limitations of these methods lie in the fact that firstly, they only allow uniform change across an image, or at most, allow partitioning the picture into an unchanged and a changed region according to a binary map. This is sub-optimal, given that introducing a new concept into the source image with only binary change capability often leads to abrupt transformations between the retained regions and the introduced regions. Secondly, they fail to make use of pre-existing foundational TTI models such as~\cite{rombach2022high, podell2023sdxl, razzhigaev2023kandinsky} which contain rich latent spaces that can conceptually bridge source and introduced context. Thirdly, these methods often take away the original ability of the model to guide generation using text while consuming large computing resources to specifically train for image editing functionality. Finally, the existing methods are strongly limited by the maximum number of exemplars they have been trained to work with (in most cases, just one). To deal with these shortcomings, we propose a general-purpose image editing approach that leverages off-the-shelf TTI models.

\section{Method}
Our goal is to improve exemplar-based image editing along two key dimensions: controllability (down to arbitrary image regions) and quality (adherence and realism). Our proposed method achieves both by enabling users to define the strength of change at each pixel via a non-binary edit map.

\textbf{Background and Setup.}
Diffusion models (DMs) are a class of generative models capable of image-to-image translation using a forward process to gradually add Gaussian noise to the image for a fixed Markov chain of $t_{ds}$ steps, and a reverse process to gradually denoise the corrupted image with a learnable model~\cite{ho2020denoising}. Denoising strength $t_{ds}$ used controls the timestep at which we start the reverse diffusion process. In this paper, we use their efficient successor, Latent Diffusion Models~\cite{podell2023sdxl} that operate on a lower dimensional representation (latent space), using latent encoders and decoders to translate the input image and output latent across pixel-latent manifolds.

\begin{figure}
\centering
\includegraphics[width=0.45\textwidth]{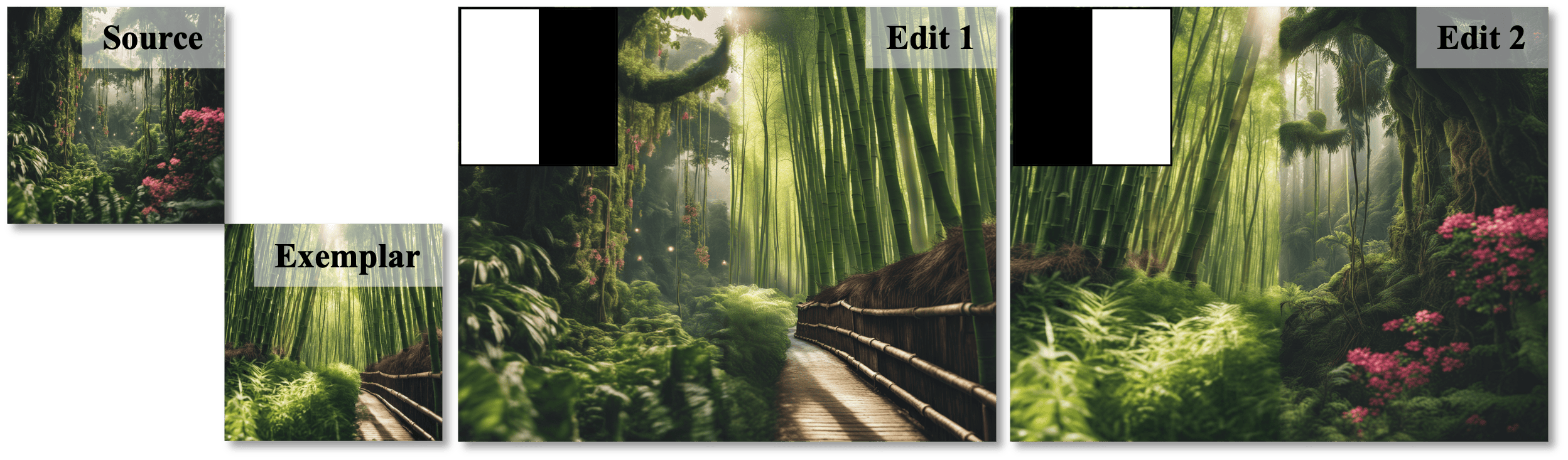} 
\caption{\textbf{Illustration of transition artifacts in the naive solution}, generating unrealistic edits.}
\label{fig:binaryblend}
\end{figure}

\textbf{A Naive Solution.}  A naive solution for this task is to use a binary mask to directly replace the masked pixels in the source image with the exemplar condition. This ``surgical latent'' is then allowed to naturally denoise through the reverse diffusion process in a TTI model for projecting the latent into the real image distribution. However, applying to test images, we find that the generated edit is far from satisfactory. There exist obvious transition artifacts across the edit boundary, making the
result look extremely unnatural, as seen in Fig.~\ref{fig:binaryblend}. We argue that this is because typical image generation models have not been trained with image editing objectives. Hence, with this naive scheme applied during inference, the model fails to adapt the exemplar to the source nor successfully hallucinate how these latents can co-exist realistically when placed next to each other by latent surgery.

\begin{figure*}[!t]
\centering
\includegraphics[width=0.86\textwidth]{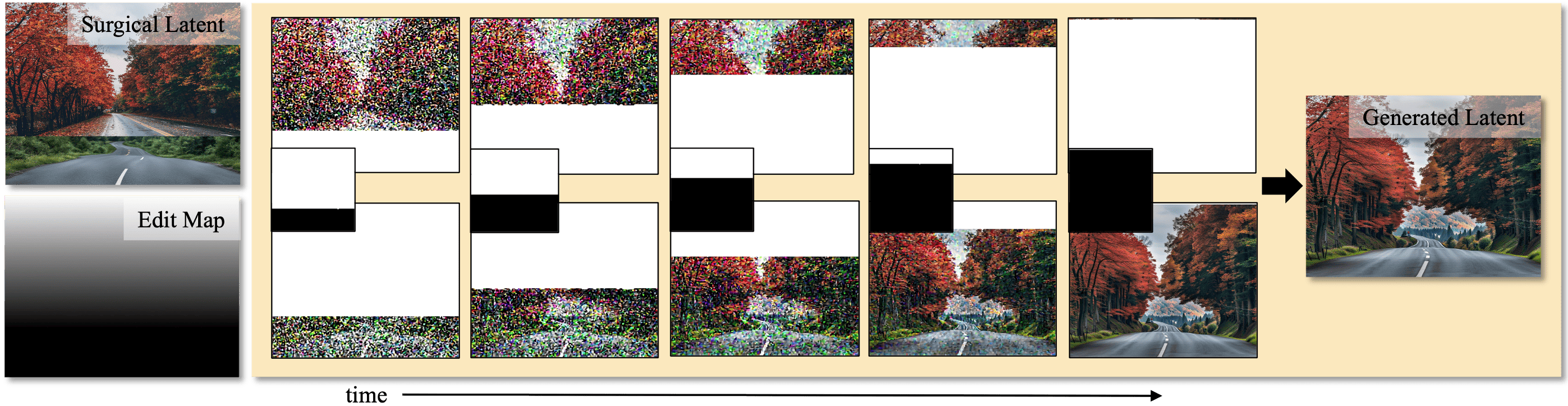} 
\caption{\textbf{Visualization of Algorithm~\ref{alg:inference} Line~\ref{line14} over time.} Top:
$z_1^{t} \odot mask_t$, regions copied from a noised version of the input. Bottom: $z^{t+1}_{mix} \odot (1 - mask_t)$, residue regions copied from the U-Net output in previous step. Note how the shifting mask at each timestep controls the inference process - darker the corresponding region in the edit map, the earlier it is copied from the residue. For ease of understanding, images are shown in the pixel space instead of the latent space.}
\label{fig:arch}
\end{figure*}

\textbf{Traversing to the Real Image Manifold.} Our key hypothesis is as follows: Given a surgical latent $ z_{surgical}$, created between the source and exemplar images, we first perturb it with Gaussian noise to induce smoothing out of transition artifacts across the edit. We start by sampling from $z_{surgical}(t_{ds}) \sim N( z_{surgical}, \sigma^2(t_{ds})I)$~\cite{song2020denoising} starting at any denoising strength $t_{ds} \in (0,T)$ (lower the $t_{ds}$, later its inference begins). This is followed by progressively removing the noise by reverse diffusion. This process allows mapping data from the noised surgical latent distribution to a latent state $z_{surgical}(0)$ (abbreviated as $z(0)$ hereafter) in the manifold of realistic images. 

We aim to bound expected squared distance between the initial latent $z_{surgical}(t_{ds})$  and its final state  $z(0)$, to estimate the adherence between them. Supposing the latent decoder model is $K$-Lipschitz, we have that similar latent states lead to similar enough decoded images. Formally, $\|z_{surgical} - z(0)  \|\leq K \|x_{surgical} - x(0)  \|$, where $x_{surgical}$ is the decoded surgical image with copy-paste artifacts (see edit results in Fig.~\ref{fig:binaryblend}) and $x(0)$ is the generated realistic edit. In this paper, we present an algorithm for progressive editing to enforce constrained traversal between a surgical latent and the generated latent (with desired edit) lying in the real image distribution. This translates to constrained change in the image space, ensuring high adherence between the resulting edit and original images. 
Following \cite{yang2023lipschitz}, we assume that a finite value $B$ exists s.t $\forall z, B = \text{sup} \, ||s_\theta(z, \sigma))||^2$ where $s_\theta$ is the model learnt during DM training. We present the following lemma:

\emph{Lemma 1:} 
For all $p \in (0,1)$, we have,
\small 
\begin{equation}
\begin{aligned}
\mathbb{P}\Big(\mathbb{E}\big[\|z(0) - &z_{surgical}(t_{ds})\|^2\big] 
\leq \sigma^4(t_{ds})B + \sigma^2(t_{ds}) \big(k \\
&+ 2 \sqrt{-k\log p} - 2 \log p \big)\Big) \geq (1-p)
\end{aligned}
\end{equation}
\normalsize 
where $z_{surgical}$ has $k$ degrees of freedom. For proof of lemma 1, please refer to the Appendix. It provides an upper bound on the change from the unrealistic surgical latent to the realistic edited image as a function of the denoising strength. We find this to be a two-way street, defining how the expected difference and hence adherence between the surgical latent and the generated latent (measured by their distance in the latent space) can be controlled by $t_{ds}$, while also allowing intuition on the choice of $t_{ds}$ based on the expected difference between the latents. If the surgical latent is an unrealistic composition, e.g, created between images taken in summer and winter, then even the nearest edited image in the real image manifold could be at a much larger distance, indicating that we must increase the denoising strength to allow farther traversal between the surgical latent and real image distributions. In this case, the distance between the latents increases as denoising strength is increased, encouraging reduced adherence to the exemplar while allowing added flexibility to achieve realism.

\textbf{Local Strength Control Using Edit Maps.}
Inspired by \cite{levin2023differential}, our method takes this controlled traversal a step further. Instead of a uniform scalar \(t_{ds}\) applied to the entire edited region, our algorithm works with local strength control, defining \(t_{ds}\) for every pixel in the edit, to enable greater flexibility and finer-grained control over the output. In the previous example, when progressively editing between summer and winter images such that the scene transitions between seasons across the width of the image, selectively modifying different regions starting at different timesteps during the diffusion inference can control spatial fidelity to the original images. Moving from the summer side to the winter side, pixels are progressively allowed more denoising strength, thus creating a smooth transition across concepts. However, to make our approach compatible with off-the-shelf DMs that only enable a scalar denoising strength setting, we design our method to instead implement this control using a non-binary `edit map' matrix for specifying the quantity of change at each spatial location. 

\begin{algorithm}[h]
\caption{Progressive Image Editing}\label{alg:inference}
\begin{algorithmic}[1]
\small
\State \textbf{Input:} $x_1$ (source image), $x_2$ (exemplar), $\mu$ (edit map), $t_{ds}^{max}=T$ (maximum denoising strength), $p= $``'' (prompt)
\State \textbf{Output:} $\hat{x}$
\Procedure{Inference}{$x_1, x_2, \mu, T, p$}
    \State $z_1^{init} \gets \text{ldm\_encode}(x_1)$
    \State $z_2^{init} \gets \text{ldm\_encode}(x_2)$
    \State $\mu_{d} \gets \text{down\_sample}(\mu)$
    \State $z_1^T \gets \text{add\_noise}(z_1^{init}, T)$
    \State $z_2^T \gets \text{add\_noise}(z_2^{init}, T)$
    \State $mask_T \gets \mu_{d} > 0$
    \State $z_{mix}^T \gets z_1^{T} \odot mask_T + z_2^{T} \odot (1 - mask_T)$\label{line10}
    \State $z_{mix}^T \gets \text{denoise}(z_{mix}^T, p, T)$
    \For{$t = T-1$ \textbf{to} $0$}
        \State $z_1^{t} \gets \text{add\_noise}(z_1^{init}, t)$
        \State $mask_t \gets \mu_{d} > (T - t)/T$
        \State $z_{mix}^t \gets z_1^{t} \odot mask_t + z_{mix}^{t+1} \odot (1 - mask_t)$\label{line14}
        \State $z_{mix}^t \gets \text{denoise}(z_{mix}^t, p, t)$
    \EndFor
    \State $\hat{x} \gets \text{ldm\_decode}(z^0_{mix})$
    \State \Return $\hat{x}$
\EndProcedure
\end{algorithmic}
\end{algorithm}

\textbf{Algorithm for Progressive Image Editing.}
Our method takes over the inference algorithm during reverse diffusion steps(Algorithm~\ref{alg:inference}). We start by encoding the source and exemplar images to the latent space ($z_1^{init}$ and $z_2^{init}$ respectively), while the edit map is downsampled to the spatial dimensions of the latent tensors. As examined by~\cite{levin2023differential}, we find latent encoders used in~\cite{podell2023sdxl, razzhigaev2023kandinsky} typically encode pixels to the same relative positions. This allows the down-sampled map ($\mu_{d}$) to align with positions of pixels in the latent representation. The highest edit strength $t_{ds}^{max}$ is set by the maximum number of inference steps. We start with the surgical latent at line~\ref{line10}. At each timestep $t$ in the denoising loop, latents are noised to the current timestep and the regions taken directly from the noisy source decreases, while more of the latent is composed of pixels copied from the denoising output in previous steps. This transition is gradual and controlled by a mask of all points lower than the current threshold determined by normalized timestep count (depicted in Fig.~\ref{fig:arch}). This linearly shifting mask determines when each region begins inference: pixels that start inference earlier (in the darker regions) have lesser ``memory'' of the source concept and attain more flexibility to adapt to the exemplar concept, while it works the other way around the later you start inference. This gradual exposure is key to how the model can generate intermediate features that naturally connect and transition across the edit seamlessly. Because DMs are not trained on intermediate images with holes, this process mimics its training distribution and gives it advance knowledge of the content in lighter regions. After the loop, the U-net output $z^0_{mix}$ is
decoded to the pixel space as generated edit $\hat{x}$. Our method can be extended to an arbitrary number of exemplars by replacing lines~\ref{line10} and~\ref{line14} in Algorithm~\ref{alg:inference} with nested functions for each new exemplar. See the Appendix for the explicit algorithm.

\begin{figure*}
\centering
\includegraphics[width=0.99\textwidth, height=1.0\textheight]{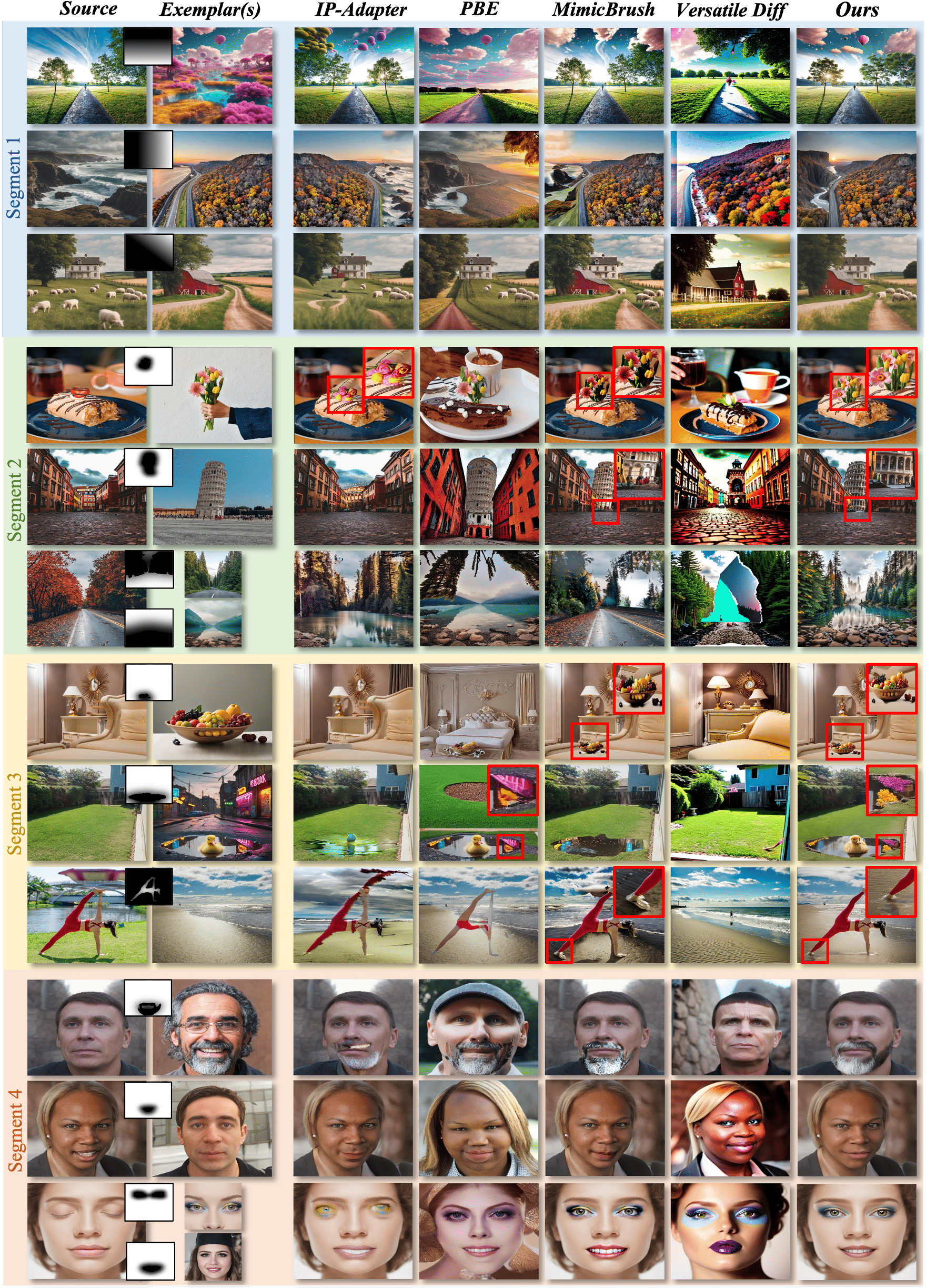} 
\caption{\textbf{Qualitative comparisons.} Our method can create realistic edits with high source and exemplar consistency.}
\label{fig:qualcomp}
\end{figure*}

\textbf{Spatially Adjusting Noise.} A simple baseline involves spatially adjusting the added noise magnitude to generate similar edits to our method.  However, we find that multiplying the added noise by an edit map tends to cause convergence to random single-color images. This is intuitive since the model has been trained to handle a specified noise distribution, which this scheme disrupts.

\textbf{Implementation Details.}
For experiments, we use off-the-shelf SDXL~\cite{podell2023sdxl} for denoising. However, the algorithm can be generalized across any DM (See the Appendix). We do not assume anything about source of the edit maps and find that they can be easily generated by operations like growing and blurring, or simple histogram transformations on binary masks created using tools like Language Segment-Anything, user interaction, automatic depth maps~\cite{Miangoleh2021Boosting} etc. Unless stated otherwise, we keep text prompts empty ($p= $``'') for all experiments.

\captionsetup[subtable]{subrefformat=simple,labelformat=simple}
\renewcommand{\thesubtable}{\thetable\alph{subtable}}

\begin{table}[t]
    \centering
    \begin{minipage}[t]{0.05\linewidth}
        \raggedleft (a)
    \end{minipage}%
    \begin{minipage}[t]{0.95\linewidth}
        \centering
        \begin{subtable}[t]{\linewidth}
            \label{metrics}
            \centering
            \resizebox{\linewidth}{!}{ 
                \begin{tabular}{lcc}
                    \hline
                    Method        & CLIP-I Score ($\uparrow$) & FID ($\downarrow$)\\ \hline \vspace{1pt}
                    Versatile Diffusion~\cite{xu2023versatile}  & 68.79 & 18.174     \\ \vspace{1pt}
                    Paint-By-Example (PBE)~\cite{yang2023paint}     &       78.67  &   8.543   \\ \vspace{1pt}
                    IP-Adapter~\cite{ye2023ip}      &   81.32   &       7.958      \\ \vspace{1pt}
                    MimicBrush~\cite{chen2024zero}      &     84.95   &      7.915      \\ \hline \vspace{1pt}
                    PIXELS (ours)        &   \textbf{91.71}     &     \textbf{5.412}    \\ \hline
                \end{tabular}
            }
        \end{subtable}
    \end{minipage}
    \vskip\baselineskip
    \begin{minipage}[t]{0.05\linewidth}
        \raggedleft (b)
    \end{minipage}%
    \begin{minipage}[t]{0.95\linewidth}
        \centering
        \begin{subtable}[t]{\linewidth}
            \label{userstudy}
            \centering
            \resizebox{\linewidth}{!}{ 
                \begin{tabular}{lcc}
                \hline
                    Method              & Adherence                       & Realism  \\ \hline \vspace{1pt}
                    Versatile Diffusion~\cite{xu2023versatile} & 4.93 &   4.03    \\ \vspace{1pt}
                    Paint-By-Example (PBE)~\cite{yang2023paint}    &      3.07                        &    3.10        \\ \vspace{1pt}
                    IP-Adapter~\cite{ye2023ip}          &      3.30                         &    3.50        \\ \vspace{1pt}
                    MimicBrush~\cite{chen2024zero}        &        2.27                     &      2.97      \\ \hline \vspace{1pt}
                    PIXELS (ours)       &     \textbf{1.40}                         &   \textbf{1.40}         \\ \hline
                \end{tabular}
            }
        \end{subtable}
    \end{minipage}
    \caption{\textbf{(a) Quantitative comparison across methods.} We evaluate the generated image quality using FID score and semantic consistency to the exemplar through CLIP-I score. \textbf{(b) User study results.} Average ranking score for semantic adherence and visual realism. $1$ is the best, $5$ is the worst.}
\end{table}

\section{Results}
We start by comparing to existing works both qualitatively and quantitatively. Next, we demonstrate how our pipeline is compatible with multi-modal prompts for editing. 

\textbf{Evaluation on Exemplar-Driven Editing.} We select different categories of edits to evaluate our method. As seen in Fig.~\ref{fig:qualcomp}, our method can deal with edits across different topics and domains. Segment 1 shows examples of scene composition, useful for new image creation where the user wants to perform general edits to compose new scenes. Segment 2 illustrates the application of semantic edits with in-the-wild images. It should be noticed how our method generates seamless results by hallucinating realistic interactions between the edited-concept and the original image. This is made possible by the shifting mask which allows creating river banks and imagining reflection from the trees near the edit boundary while encouraging high source and exemplar fidelity elsewhere. Contemporaries such as PBE~\cite{yang2023paint}, IP-Adapter~\cite{ye2023ip}, and Versatile Diffusion~\cite{xu2023versatile} do not guarantee fidelity or realism for editing in-the-wild images. MimicBrush~\cite{chen2024zero} generates more realistic results, but we still observe undesired changes to the exemplar and background. Segment 3 illustrates foreground-inpainting. In the third example in this segment, we show a variation to outpaint the background instead of foreground-inpainting and create realistic interactions between the human and the beach, proving our strong generalization ability compared to prior works. In the last segment, we show comparisons in practical applications like face edits, and makeup look curation from inspiration exemplars. For visualizations from contemporaries in more-than-one exemplar setting, we edit iteratively to incorporate each new exemplar. Our method, on the other hand, shows significant superiorities by being able to perform the edit in a single pass, with as little as $0.04\%$ inference memory overhead on the original SDXL. More results in the Appendix.

For fair quantitative comparison, we sample 3000 pairs of random images from Imagenet's validation set~\cite{russakovsky2015imagenet} as inputs to all methods, with a randomly chosen edit map from our database. Since we are the first method to allow non-binary edit maps, we test other methods with the binarized version of the map. To measure the aspects of photo-realism and edit fidelity (to the source and exemplar) independently, we use the FID score~\cite{heusel2017gans}, which is widely used to evaluate realism in generated images and the CLIP-I score~\cite{radford2021learning}, measuring similarity between edited region and the exemplar. Table~\subref{metrics}a presents quantitative comparison results. Our approach achieves the best performance on both metrics, verifying that it can not only generate high-quality edits but also maintain high fidelity to the original images.

\textbf{User Study.} As quantitative metrics do not fully represent human preferences, we conduct a user study. We let 26 participants assess and rank generated results on 30 groups of samples, with each group containing the source, exemplar(s), edit map(s) and anonymized outputs from all the methods (presented in a random order). Users were asked to rank the outputs from 1 to 5 (1 being best, 5 is worst) on the criterion of adherence and realism. Adherence refers to the ability to preserve the identity of the source and exemplars according to the edit map. Realism considers if the output looks like a natural and seamless edit of good quality. These aspects are evaluated independently. Results are listed in Table~\subref{userstudy}b; our method earns significantly more preferences.

\begin{figure*}[!ht]
\centering
\includegraphics[width=0.95\textwidth]{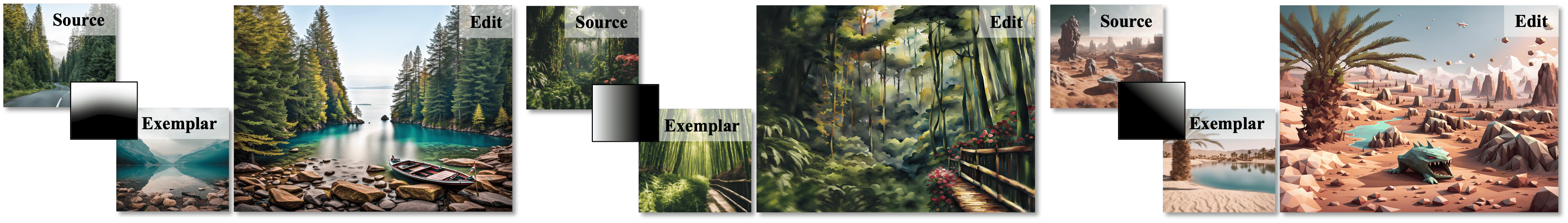} 
\caption{\textbf{Results with multi-modal prompts.} Text Prompts: “boat”, “watercolor style”, “low poly aesthetic, big monster”.}
\label{fig:textcontrol}
\end{figure*}

\begin{figure*}[!h]
\centering
\includegraphics[width=0.95\textwidth]{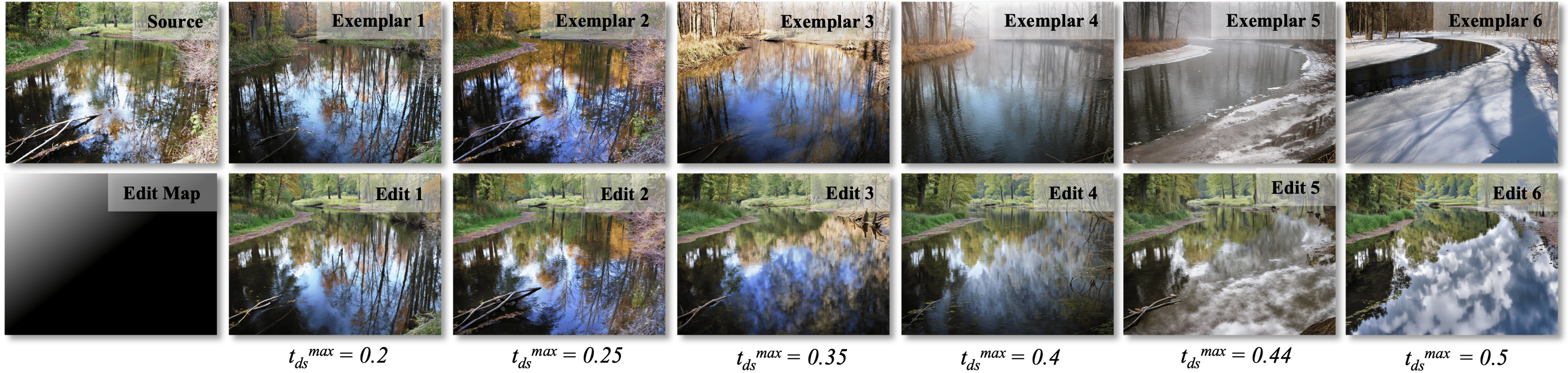} 
\caption{\textbf{Trade-off between adherence and realism as source and exemplar images grow further in the latent space (Euclidean distance)}. To reach the minimal realism score, we keep increasing $t_{ds}^{max}$ from left to right as latent distance increases (Lemma 1), causing the edits to show lower fidelity to the exemplar while increasing hallucination to create a realistic scene.}
\label{fig:tds}
\end{figure*}

\textbf{Bringing Back Text Control.} In most existing models for this task, the original text-to-image ability is lost. However, with our proposed algorithm, we can generate edits with multi-modal prompts (exemplars and/or text guidance) since we do not disrupt the original TTI model with training/ fine-tuning. Instead of using empty text prompts like we have been doing, 
we can use additional language guidance to generate more diverse edits. As seen in Fig.~\ref{fig:textcontrol}, we can edit attributes, change the generation style or do both during denoising by adding simple text descriptions.

\begin{figure}
\centering
\includegraphics[width=0.42\textwidth]{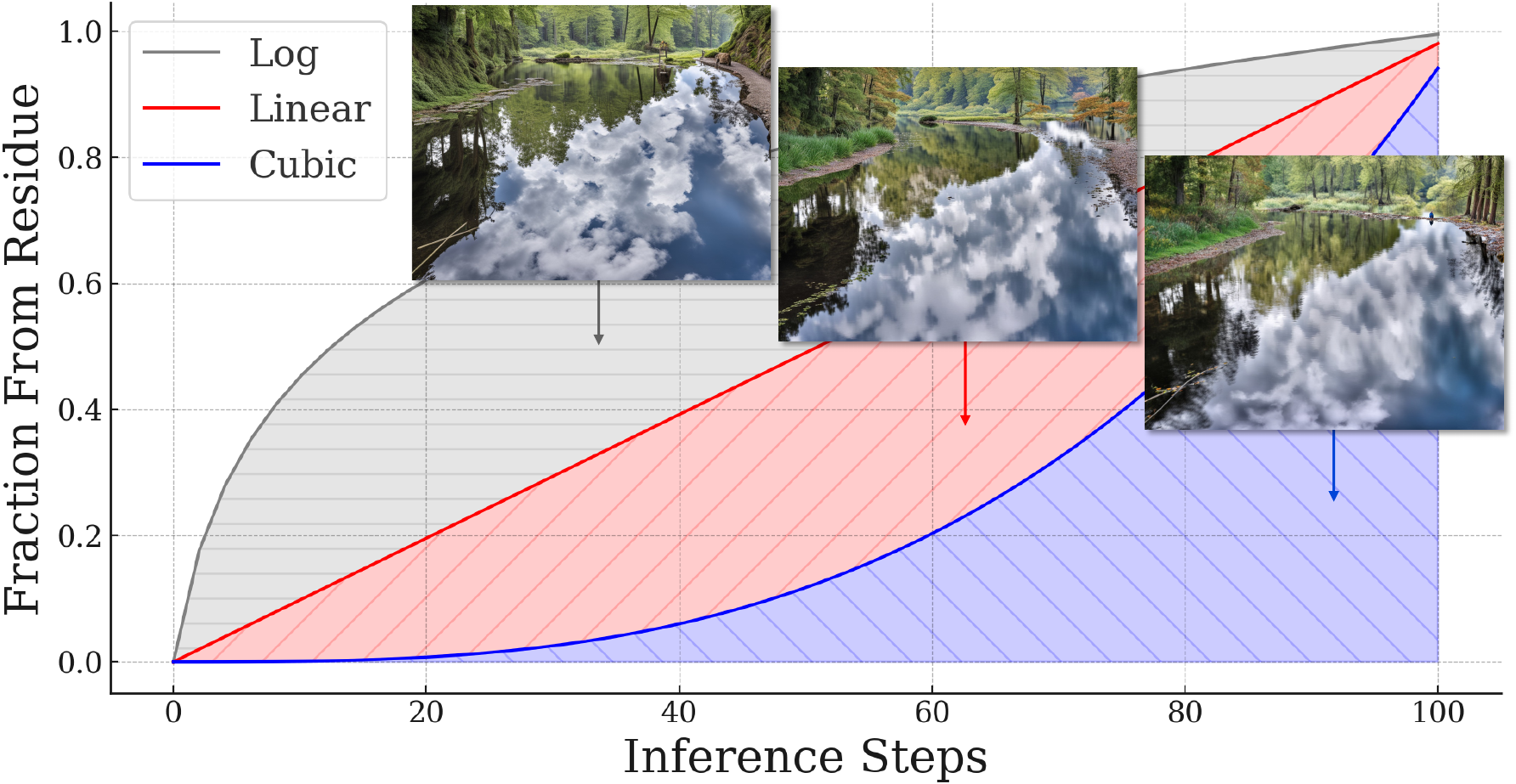} 
\caption{\textbf{Thresholding Strategies.} Setting various thresholds allow users to finetune their edits at a given strength.}
\label{fig:threshold}
\end{figure}

\section{Ablation Studies}
\textbf{Analyzing the impact of $t_{ds}$.} We note that for trained diffusion models, there exists an inherent trade-off between adherence and realism metrics with changing denoising strength. Given a source image and increasingly distant exemplars in the latent space (left to right), we aim to create edits that evaluate to at least a minimum realism score~\cite{gu2020giqa}. As visualized in Fig.~\ref{fig:tds}, for the same edit map, we increase $t_{ds}^{max}$ to maintain realism in the edit as we go from left to right, allowing the model to exert more hallucination to imagine realistic compositions while becoming less and less adherent to the exemplar. Notice how for exemplar 6, the model leverages the high strength to hallucinate the snowy bank as a cloudy sky for the edit to look natural with high probability (lemma 1), since the surgical latent has an unrealistic composition with winter and summer scenes co-existing in the same image. More details in the Appendix.

\hspace{10pt}\textbf{Mask Thresholding Strategies.} Selecting $t_{ds}$ provides a coarse mechanism for controlling the amount of noise perturbation and hence, the edit strength. We introduce an additional fine adjustment parameter that allows users to precisely modulate the generation results through the choice of an appropriate threshold function, as explored in Fig.~\ref{fig:threshold}. At a constant denoising strength, this effectively decides the relative amount of time each noised region corresponding to a brightness level from the edit map spends inside the inference loop (shown as Area Under Curve). Besides the default linear threshold that works well across edits and is hence used for all experiments in this paper, we find that users can switch to Log control to encourage higher fractions of the latent $z_{mix}^t$ to be copied from the U-net residue (hence, more flexibility to hallucinate), or use Cubic control (more Area Above Curve) to rally for copying noisy latent regions from the original image for more steps (hence higher adherence). See the Appendix for more visualizations and details.

\section{Limitations \& Potential Impacts}
Our method can work with in-the-wild images, allowing for easy photo manipulation and creating content with negative societal impacts. On the other hand, this enables
everyday users with little to no artistic expertise to create realistic edits with ease, lowering the barrier to entry for visual content creation. To address this, we will specify permissible uses of the method with appropriate licenses during code release. 

Despite the effectiveness of our method, the user creates the edit map for now, using manually drawn masks, depth, or segmentation maps. In the future, tools for automating map generation can help in widespread adoption of our method.

\section{Conclusion}
We introduced a novel solution to the exemplar-driven editing task that allows region-wise control over the editing process and demonstrated its superiority over existing baselines. Our method requires no training/fine-tuning, has minimal overhead, can be used with in-the-wild images as well as text prompts, and can work with an arbitrary number of exemplars. This work hopes to bring new inspiration for the community to explore advanced solutions designed by leveraging existing large models for a more sustainable future.

\appendix
\begin{center}
    {\LARGE\bfseries Appendix}
\end{center}

In this Appendix, we provide more details on the experiments, methods and ablations presented in the main paper. Section $1$ derives the proof for Lemma 1 presented in the main manuscript; Section $2$ provides the explicit algorithm for extending our method to an arbitrary number of exemplars and delineates how we can offer single-pass as well iterative editing settings with ease. Section $3$ presents additional results to visualize edits created by our method; while Section $4$ specifies more details on the denoising strength ablation presented in the main paper, and highlights how this can effect generation quality across different settings. In Section $5$, we also extend the discussion on experiments from the main paper that show fine adjustments possible to an edit when using different mask thresholds. Section $6$ highlights details on memory and inference time overheads of introducing our algorithm to the unaltered base model. Finally, we provide a discussion on extending our method to other diffusion models in Section $7$.

\section{1. Proof of Lemma 1}
\subsection{1.1 Setup}

We consider a latent diffusion model, operating in a pretrained, learned (and fixed) latent space of an autoencoder.
Given a surgical latent $ z_{surgical}$, created between the source and reference images using a binary mask ($mask \odot img_{source} + (1 - mask) \odot img_{exemplar}$), we first perturb it with Gaussian noise to induce smoothing out of transition artifacts across the edit. We start by sampling from $z_{surgical}(t_{ds}) \sim N( z_{surgical}, \sigma^2(t_{ds})I)$~\cite{song2020denoising} starting at any denoising strength $t_{ds} \in (0,T)$ (lower the $t_{ds}$, lesser the noise perturbation). This is followed by progressively removing the noise by reverse diffusion. This process allows mapping data from the noised surgical latent distribution to a latent state in to the manifold of realistic images denoted by $z_{surgical}(0)$. 

We aim to bound expected squared distance between the initial latent $z_{surgical}(t_{ds})$ (dubbed $z(t_{ds})$ hereafter)  and its final state  $z_{surgical}(0)$ (abbreviated as $z(0)$), to estimate the adherence between them. Supposing the latent decoder model is $K$-Lipschitz, we have that similar latent states lead to similar enough decoded images. Formally, $\|z_{surgical} - z(0)  \|\leq K \|x_{surgical} - x(0)  \|$, where $x_{surgical}$ is the decoded surgical image with copy-paste artifacts (refer to edit results in Fig. 2 from the main manuscript) and $x(0)$ is the generated edit. In the main paper, we presented an algorithm for progressive editing to enforce constrained traversal between a surgical latent and generated latent (with desired edit) lying in the real image distribution. This translates to constrained change in the image space, ensuring high adherence between the edit and original images. Here, we provide proof of the presented Lemma 1.

\subsection{1.2 Perturbing Latent Space with Noise}
Authors in \cite{song2020score} construct a diffusion process \( \{z(t)\}_{t=0}^{T}\)
indexed by a continuous time variable \(t \in [0, T]\), such that \(z(0) \sim p_0\), for which we have a dataset of i.i.d. samples, and \(z(T) \sim p_{T}\), for which we have a tractable form to generate samples efficiently. The stochastic differential equation for latent space dynamics:
\begin{equation}
dz = f(z, t) \, dt + G(z, t) \, dw_t
\end{equation}
where:
\begin{itemize}
    \item \( f(\cdot, t) \): \(\mathbb{R}^d \rightarrow \mathbb{R}^d\) is a vector valued function called the drift coefficient of z(t); the drift function aims to reverse the diffusion effects, guiding the latent states toward the initial or less noisy states.
    \item \( G(\cdot, t) \): \(\mathbb{R}^d \rightarrow \mathbb{R}^d\) is a matrix function known as
the diffusion coefficient of \( z(t) \) that models the standard deviation of the noise process, accounting for varying noise levels across different latent dimensions. 
    \item \( w_t \) is the standard Wiener process (a.k.a., Brownian motion).
\end{itemize}
The drift component is deterministic (looking at the formulation of ODEs in \cite{song2020score}), but the diffusion component is stochastic due to the standard Wiener process.

\subsection{1.3 Generating Edits by Denoising}
By starting from noisy surgical samples of \( z(t_{ds}) \sim p_{t_{ds}}\) and reversing the process, we can obtain real edited samples \( z(0)  \sim p_0\). Results from \cite{anderson1982reverse} show that this can be performed by a reverse-time SDE when time flows backwards from \(t_{ds}\) to \(0\):
\begin{equation}
dz = [f(z, t) - G(z, t)^2 \nabla_z \, \text{log} \, p(z, \sigma)] \, dt + G(z, t) \, d\bar{w}_t
\end{equation}

where \(\bar{w}_t\) is a standard Wiener process in the reverse time, with \(\bar{w}_t\) independent of past increments of \(\bar{w}_t\), but not of future ones. 
Once the score of each marginal distribution $\nabla_z \, \text{log} \, p(z, \sigma)$ is known
for all \(t\) (estimated using a time-dependent score-based model \(s_\theta(z, \sigma\)), we can derive the reverse diffusion process from the above equation and simulate it to move from the prior surgical latent distribution \( p_{t_{ds}}\) and finally sample from real image distribution \(p_0\).

Decomposing the reverse process from \( t_{ds} \) to 0:\
\small
\begin{equation}
\begin{aligned}
z(0) &= z(t_{ds}) + \int^{t_{ds}}_0 [f(z, t) - G(z, t)^2 \nabla_z \, \text{log} \, p(z, \sigma)] \, dt \\
&+ G(z, t) \, d\bar{w}_t \\
&= z(t_{ds}) + \int^{t_{ds}}_0 [f(z, t) - G(z, t)^2 s_\theta(z, \sigma)] \, dt \\
&+ G(z, t) \, d\bar{w}_t 
\end{aligned}
\end{equation}
\normalsize

Taking the squared \( L^2 \) norm and applying expectation: 
\small
\begin{equation}
\begin{aligned}
E[\|z(0) - z(t_{ds})\|^2] 
&= E\bigg[\bigg\|\int^{t_{ds}}_0 \big(f(z, t) - G(z, t)^2 \cdot s_\theta(z, \sigma)\big) \, dt \\
&\quad + G(z, t) \, d\bar{w}_t \bigg\|^2\bigg]\\
&\leq E\left[\left\|\int^{t_{ds}}_0 f(z, t) \, dt\right\|^2\right] \\
&+ E\left[\left\|\int^{t_{ds}}_0 -G(z, t)^2 s_\theta(z, \sigma) \, dt\right\|^2\right] \\
&+ E\left[\left\|\int^{t_{ds}}_0 G(z, t) \, d\bar{w}_t\right\|^2\right] \\
&\text{\small{(triangle inequality theorem)}}
\end{aligned}
\end{equation}

\normalsize

\subsection{1.4 Bounding the Deterministic Components}

These coefficients are selected differently for the variance preserving (VP) and variance exploding (VE) formulations based on the behavior of the variance during evolution and can be chosen generally. Both specifications formalize the notion that the \(z(t_{ds})\) is increasingly noisy as we go forward in time. We follow the VE-SDE formulation for the rest of the proof, but our theory can be generalized to hold in either case. Following \cite{yang2023lipschitz}, we assume that a finite value \(B\) exists such that s.t $\forall z, \, B = \text{sup} \, ||s_\theta(z, \sigma)||^2$. Also, for simplicity as in \cite{song2020score} and in practice, we assume that \(G(\cdot)\) is a vector (instead of a \(d * d\) matrix) and does not depend on \(z(t)\), but our theory can be generalized to hold in those cases. Hence,
\small
\begin{equation}
    \begin{aligned}
    E\left[\left\|\int^{t_{ds}}_0 -G(z, t)^2 s_\theta(z, \sigma) \, dt\right\|^2\right] &\leq  B*\\
    &E\left[\left\|\int^{t_{ds}}_0 -G(z, t)^2 \, dt\right\|^2\right] \\
    \end{aligned}
\end{equation}
\normalsize

Assuming $\sigma(0) = 0$, which is typical for such SDE problems and making choice of coefficients for VE-SDE with a zero drift term \(f = 0\) and diffusion term \(= \sqrt{\frac{d[\sigma^2(t)]}{dt}}\):
\small
\begin{equation}
E\left[\left\|\int^{t_{ds}}_0 -G(z, t)^2 s_\theta(z, \sigma) \, dt\right\|^2\right] \leq B*E[\sigma^4(t_{ds})] \\
\end{equation}

\normalsize

\setcounter{algorithm}{1}
\begin{algorithm}[t]
\caption{Progressive Image Editing for More-Than-One Exemplar Setting}\label{alg:inference2}
\begin{algorithmic}[1]
\small
\State \textbf{Input:} $x_1$ (source image), $x_2$ (exemplar1), $x_3$ (exemplar2), $\mu_1$ (edit map1), $\mu_2$ (edit map2), $t_{ds}^{max} = T$ (maximum denoising strength), $p = $``'' (prompt)
\State \textbf{Output:} $\hat{x}$
\Procedure{Inference}{$x_1, x_2, x_3, \mu_1, \mu_2, T, p$}
    \State $z_1^{init} \gets \text{ldm\_encode}(x_1)$
    \State $z_2^{init} \gets \text{ldm\_encode}(x_2)$
    \State $z_3^{init} \gets \text{ldm\_encode}(x_3)$
    \State $\mu_{1d} \gets \text{down\_sample}(\mu_1)$
    \State $\mu_{2d} \gets \text{down\_sample}(\mu_2)$
    \State $z_1^T \gets \text{add\_noise}(z_1^{init}, T)$
    \State $z_2^T \gets \text{add\_noise}(z_2^{init}, T)$
    \State $z_3^T \gets \text{add\_noise}(z_3^{init}, T)$
    \State $mask_1 \gets \mu_{1d} > 0$
    \State $mask_2 \gets \mu_{2d} > 0$
    \State $z_{mix}^{T} \gets (z_1^{T'} \odot mask_1 + z_2^{T} \odot (1 - mask_1)) \odot mask_2 + z_3^{T} \odot (1 - mask_2)$\label{multiline14}
    \State $z_{mix}^T \gets \text{denoise}(z_{mix}^T, p, T)$
    \For{$t = T-1$ \textbf{to} $0$}
        \State $z_1^{t} \gets \text{add\_noise}(z_1^{init}, t)$
        \State $mask_1 \gets \mu_{1d} > (T - t)/T$
        \State $mask_2 \gets \mu_{2d} > (T - t)/T$
        \State $z_{mix}^t \gets (z_1^{t} \odot mask_1 + z_{mix}^{t+1} \odot (1 - mask_1)) \odot mask_2 + z_{mix}^{t+1} \odot (1 - mask_2)$\label{multiline18}
        \State $z_{mix}^t \gets \text{denoise}(z_{mix}^t, p, t)$
    \EndFor
    \State $\hat{x} \gets \text{ldm\_decode}(z^0_{mix})$
    \State \Return $\hat{x}$
\EndProcedure
\end{algorithmic}
\end{algorithm}

\subsection{1.6 Stochastic Integral Analysis}
The 3rd term represents the sum of a L2-norm squared scaled Wiener process over the interval \([0, t_{ds}]\). Each incremental part of the Wiener process \(d\bar{w}_t \) is normally distributed with mean \(0\) and variance \(\sigma^2(t_{ds})-\sigma^2(0)\), (formally, \(d\bar{w}_t \sim N(0, \sigma^2(t_{ds})-\sigma^2(0))\)) \cite{szabados2010elementary}; making the unit variance counterpart of this term distributed according to the central chi-squared distribution with k-degrees of freedom:

\small
\begin{equation}
    \frac{E\left[\left\|\int_0^{t_0} G(z(t), t) \, d\bar{w}_t\right\|^2\right]}{\sigma^2(t_{ds})} \sim \chi_k^2
\end{equation}
\normalsize

From \cite{laurent2000adaptive}, we define the following upper tail bound for any \(-\text{log} \, p > 0\) such that \(p \in (0, 1)\):

\small
\begin{equation}
\mathbb{P}\Bigg(\frac{E\left[\left\|\int_0^{t_0} G(z(t), t) \, d\bar{w}_t\right\|^2\right]}{\sigma^2(t_{ds})} - k \geq 2 \sqrt{-k\text{log} \, p} - 2 \, \text{log} \, p\Bigg) \, \leq \, p
\end{equation}
\normalsize

\subsection{1.7 Deriving the Probabilistic Upper Bound}
\begin{figure}[t]
\centering
\includegraphics[width=0.45\textwidth]{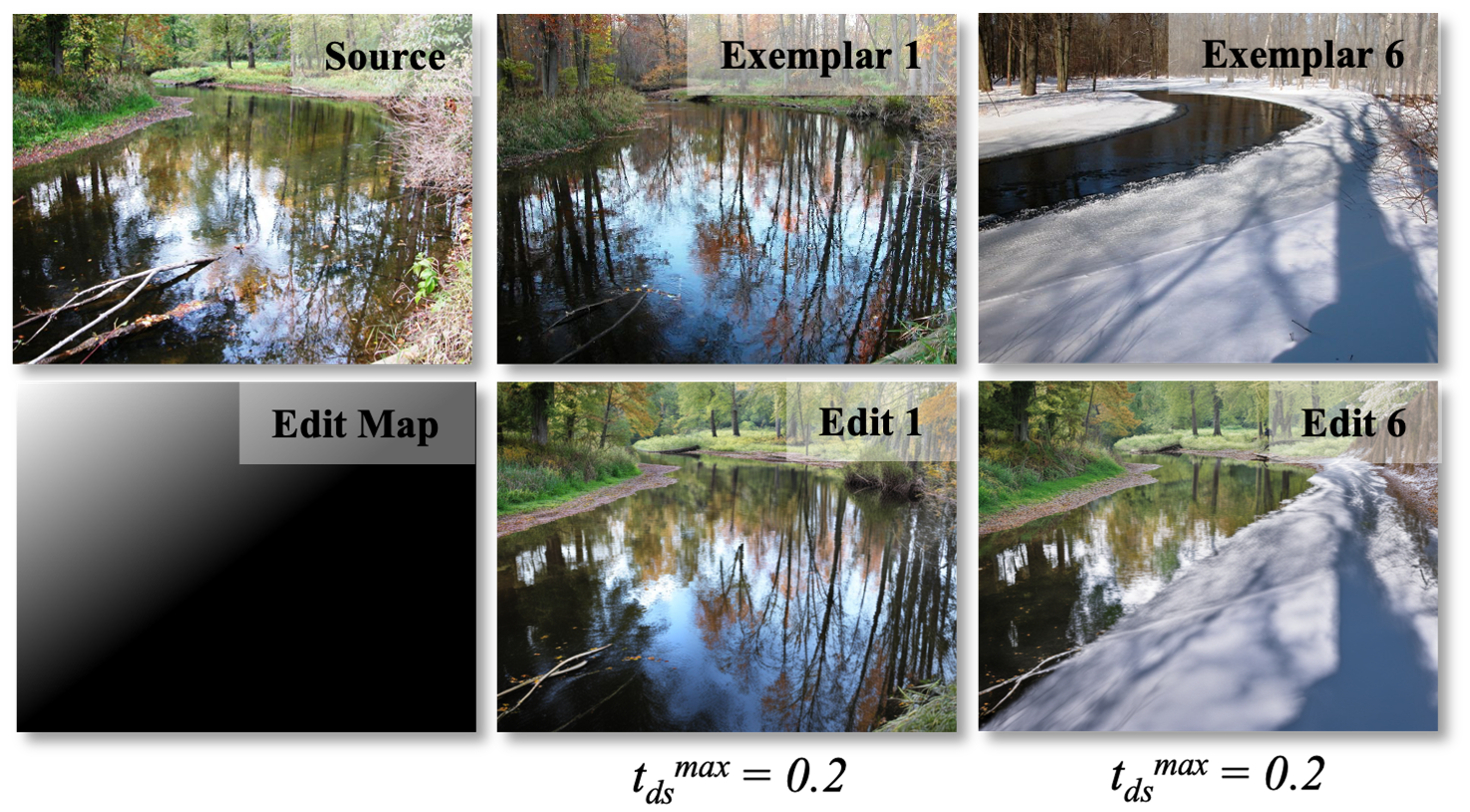} 
\caption{\textbf{Results when maintaining a constant $t_{ds}^{max}$.} This leads to unrealistic edits with exemplar $6$ when denoising strength is not increased to allow more hallucination.}
\label{fig:tdsapp}
\end{figure}

Bringing both the components together:

\small
\begin{equation}
\begin{aligned}
\mathbb{P}(E[\|z(0) - z(t_{ds})\|^2] &\leq \sigma^4(t_{ds})B + \sigma^2(t_{ds})(k \\
&+ 2 \sqrt{-k\text{log} \, p} - 2 \, \text{log} \, p)) \geq (1-p)
\end{aligned}
\end{equation}
\normalsize
This implies that for all \(p \in (0, 1)\) with probability at least \((1 - p)\), the expected squared distance between the noisy surgical latent state  \( z(t_{ds}) \) starting at any denoising strength  \( t_{ds} \) and the realistic edited latent  \( z(0) \) over the interval \( [0, t_{ds}]\) can be upper bounded as a function of the denoising strength. We find this to be a two-way street, defining how the expected difference and hence fidelity between the surgical latent and the generated latent containing the edit (measured by their distance in the latent space) can be controlled by \(t_{ds}\), whilst also allowing intuition on the choice of \(t_{ds}\) based on how large expected squared distance between the latents are. For example, if the surgical latent is an unrealistic composition, created from summertime and wintertime images ($mask \odot img_{summer} + (1-mask) \odot img_{winter})$, then even the nearest edited image in the real image manifold could be at a much larger distance. This indicates that we must increase the denoising strength to allow a farther traversal between the surgical latent and real image distributions since there must be significant hallucination permitted to create a realistic output in this case. In this case, the distance between the latents increases as we increase the denoising strength, causing reduced fidelity to the exemplar reference and increased hallucination. Refer to Fig.~\ref{fig:tdsapp} to see how not increasing the denoising strength in the $3^{rd}$ column leads to unrealistic outputs. In contrast, Fig. 6 in the main manuscript, highlights the benefits of following lemma 1 to increase the denoising strength in such cases.

\section{2. Exemplar Count Agnostic Algorithm}

In the main paper, we detail the modification of the original algorithm
to enable using an arbitrary number of exemplars for a single-pass editing workflow. We write the explicit
exemplar count agnostic algorithm in Algorithm~\ref{alg:inference2}. We show modifications allowing the use of two exemplars simultaneously for easy understanding. Lines~\ref{multiline14} and~\ref{multiline18} can be adapted as shown in the algorithm to extend the equation for incorporating as many exemplars as the user likes. Please note that we denote $\odot$ as an element-wise multiplication operation in the algorithm. We also allow for an iterative editing setup: users can incorporate one exemplar at a time using Algorithm 1 in the main paper. Generated edit $\hat{x}$ at every pass can be used as source image $x1$ for the next pass to introduce a new exemplar to the previous edit output.

\begin{figure*}[!t]
\centering
\includegraphics[width=1.0\textwidth]{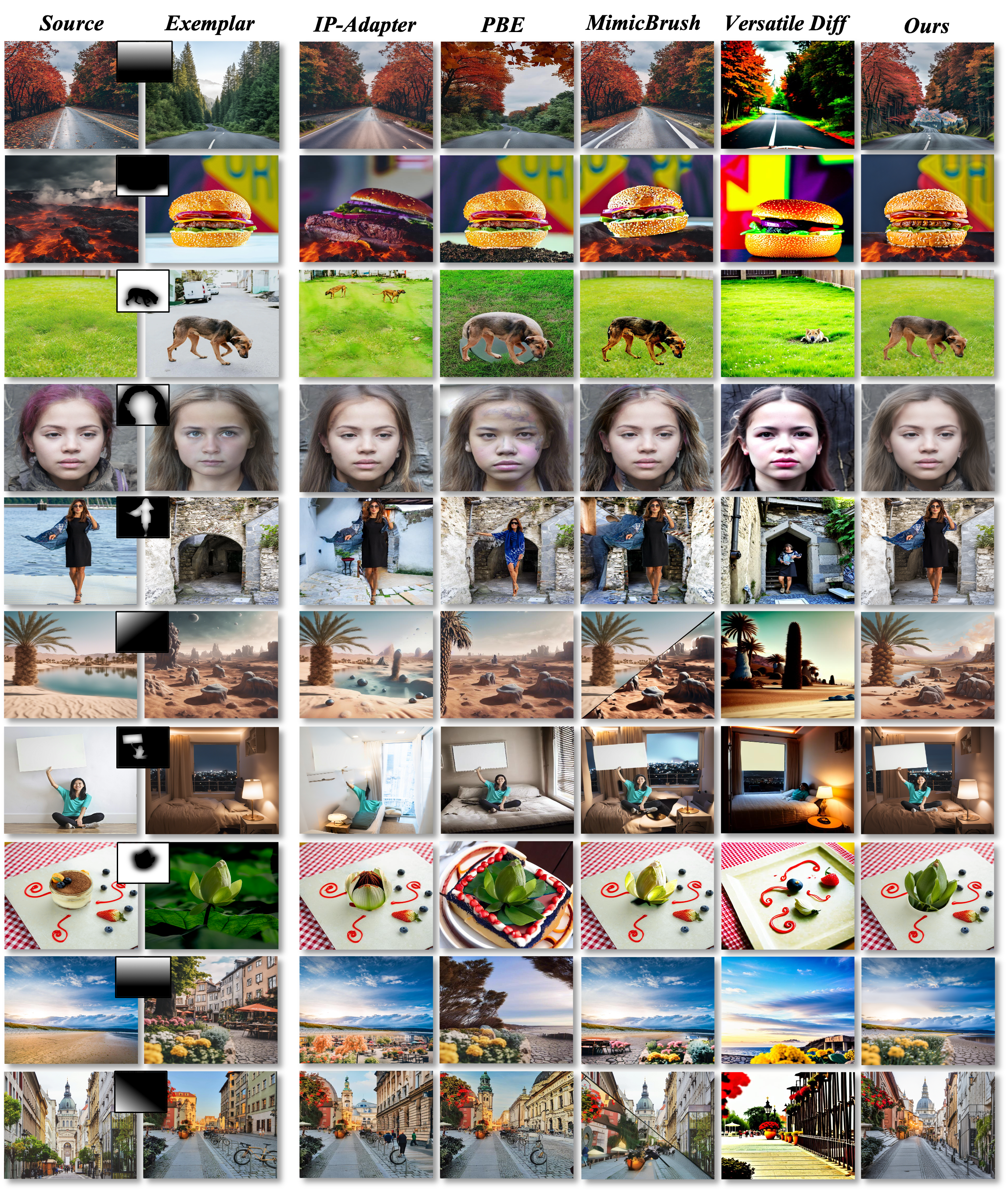} 
\caption{\textbf{Additional comparisons across all methods.} We consistently maintain better adherence to the original images while creating realistic interactions in the edit. Zoom-in for a detailed view.}
\label{fig:vizapp}
\end{figure*}

\section{3. Additional Visualizations}
In this section, we provide more results across methods in different application scenarios. Fig.~\ref{fig:vizapp} demonstrates the
ability of our method in editing any region of real as well as synthetic images. The user retains full control of the spatial location and scale they want the exemplar to appear in the edit, and have the flexibility to create an input surgical latent of their choice by manually adjusting exemplar placement. Such an example appears in row 5, where the exemplar input used across all methods contains the exemplar placed at the desired location (on the bed) and padding with zeros at the remaining locations. Our method is able to understand the objects in the exemplar images and create harmonious and consistent interactions in edited regions, as seen in the $3^{rd}$ row. Zoom in to notice the realistic interactions between the grass and the dog, enabled by our method progressively editing across them, while MimicBrush~\cite{chen2024zero} creates a copy-paste effect. 

\section{4. Technical Details on Analyzing Impact of Denoising Strength Experiment}

The main paper demonstrates the adherence-realism tradeoff during editing in Fig. 6 of the manuscript. Here, we provide more details on that experiment and refer to that figure unless specified otherwise. Note that the exemplars in that figure are arranged from left to right in increasing order of latent space distance from the source image (measured by Euclidean distance). We measured the average realism score~\cite{gu2020giqa} over all the inputs and set that as the realism threshold for evaluation. Next, for each exemplar, we start with a minimum denoising strength $t_{ds}^{max}$ of $0$ and keep increasing it till the generated edit scores equal to or above the chosen realism threshold. Fig. 6 in the main paper shows that for a given realism score, maximum denoising strength must be increased with positive correlation to the distance between the latents (i.e left to right). As exemplar $6$ and the source are found to be the most distant in latent space, an edit between them must be created with higher denoising strengths for realism. Notice how the snow is hallucinated as clouds to achieve the desired score.  To show the counter-effect of keeping a constant $t_{ds}^{max}$ irrespective of latent distances, we experiment with a constant value of $0.2$ for both exemplar $1$ and $6$. The results in Fig.~\ref{fig:tdsapp} show that this can negatively impact how realistic the generated result looks after editing.

\begin{figure*}[!ht]
\centering
\includegraphics[width=1.0\textwidth]{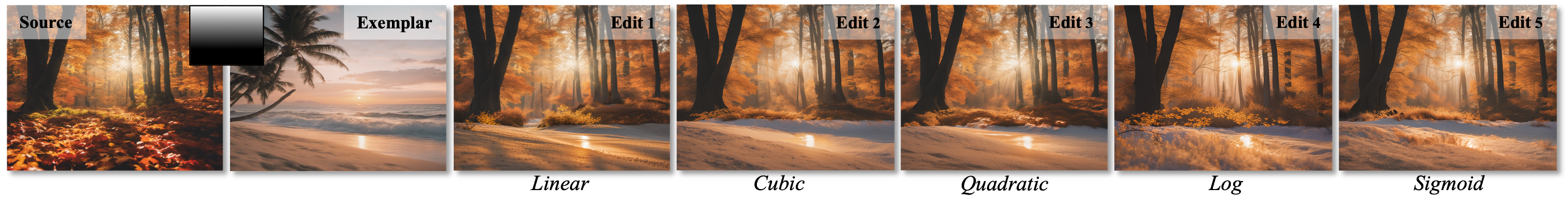} 
\caption{\textbf{Various thresholding techniques.} Our method is compatible with any thresholding function to support a wide range of editing controls. All threshold types that are examined maintain adherence to the original images according to the edit map.}
\label{fig:thresholdvar}
\end{figure*}

\begin{figure}[!t]
\centering
\includegraphics[width=0.4\textwidth]{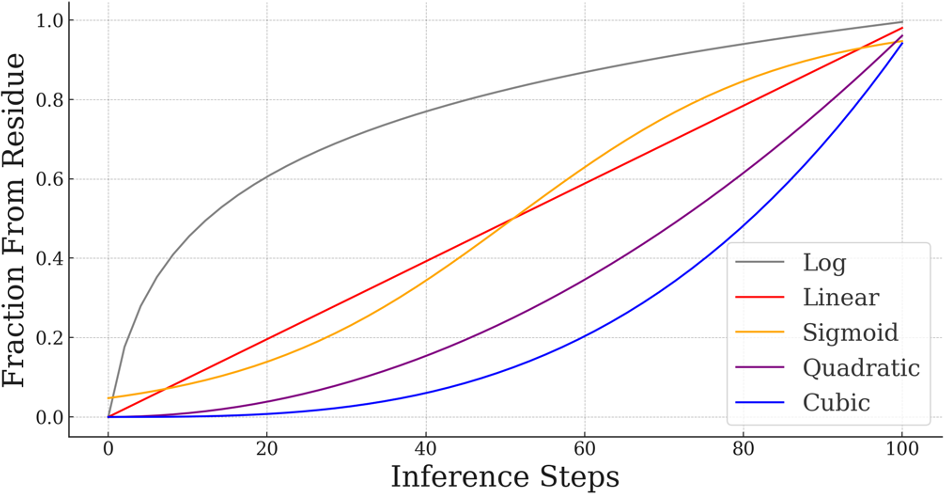} 
\caption{\textbf{Impact of choosing different thresholds} at a constant denoising strength. As Area Under Curve (AUC) increases, more fractions of the latent are copied from the residue, allowing more hallucination and lesser fidelity.}
\label{fig:thresholdvar2}
\end{figure}

\section{5. Choice of Map Thresholds}
For experiments in the main manuscript as well as Appendix, we always use the Linear Thresholding function. However, we offer additional choices for users to experiment with even beyond the Linear, Log and Cubic strategies introduced in the main paper, such as Sigmoid and Quadratic control. To explore the effect of using different thresholds at a given constant denoising strength, we experiment with the same input group while switching between each thresholding function and visualize the results in Fig.~\ref{fig:thresholdvar}. In line with our intuition from the main paper, we find that the edited region undergoes maximum deviation from the exemplar (low adherence) when using Log thresholds, while still looking realistic. On the other hand, when operating with Cubic or Quadratic thresholds, the edited region holds strong fidelity to the exemplar while looking a little less seamless at the transition boundary. This is expected, given that lower fractions of $z_{mix}^t$ are copied from the U-net residue in this case (Fig.~\ref{fig:thresholdvar2}), responsible for feature hallucination and seamless blending. Instead, copying more regions from the noisy original latent (represented by more Area Above Curve) encourages more adherence to the original exemplar and lesser possible hallucination, leading to an unrealistic transition. Sigmoid thresholds perform similarly to the Linear case, as is expected from Fig.~\ref{fig:thresholdvar2}. 
This calls back to the adherence-realism trade-off we discussed earlier. Notice how this offers a fine-tuning knob compared to the coarse denoising strength setting discussed in the earlier section. In interactive settings, we can start with linear thresholds and create edits for the user to decide if the result should have higher adherence or realism. Based on their preference, we can choose a reasonable threshold with lesser/more AUC respectively to be used for their task. Although different edits could work best with a different threshold functions, we empirically find that the default Linear threshold works well across all edit domains. Nevertheless, we define the thresholding functions mathematically below for the user to choose from:

\emph{Linear Threshold}: \small \texttt{torch.arange(len(t)) / len(t)} \normalsize

\emph{Cubic Threshold}: \small \texttt{(torch.arange(len(t)) / len(t))** 3} \normalsize

\emph{Quadratic Threshold}: \small \texttt{(torch.arange(len(t)) / len(t))** 2} \normalsize

\emph{Log Threshold}: \small \texttt{torch.log1p(torch.arange(len(t))
) / torch.log1p(torch.tensor(len(t))} \normalsize

\emph{Sigmoid Threshold}: \small\texttt{torch.sigmoid(torch.arange
(len(t)) / len(t) * 6 - 3)} \normalsize

\section{6. Overheads on Memory Consumption \& Inference Time}
We measured the inference memory consumption of Stable
DiffusionXL’s img2img~\cite{podell2023sdxl} with and without our framework. The overhead of using our method by changing the inference loop is less than $7$MB
$(0.04\%)$. Additionally, we find that the inference time overhead under the same settings compared to the original SDXL model is only around $3.86\%$, averaged over $100$ different runs.

\section{7. Extending to Other Diffusion Models}

In the main manuscript, we present our algorithm and apply it to Stable
DiffusionXL~\cite{podell2023sdxl}. However, our framework
can be generalized to other off-the-shelf diffusion models with ease. Despite their differences from the SDXL
model, our algorithm can be applied by similarly adapting Stable Diffusion 2.1~\cite{rombach2022high} and Kandinsky's~\cite{razzhigaev2023kandinsky} inference loops, without any modifications to the proposed algorithm.

\bibliography{aaai25}

\end{document}